\documentclass[runningheads]{llncs}
\usepackage{graphicx}
\usepackage{amssymb}
\usepackage{cite}
\begin{document}
\title{
A New Local Transformation Module for Few-shot Segmentation
}
%
%
\author{Yuwei Yang, Fanman Meng, Hongliang Li,
Qingbo Wu, \\Xiaolong Xu \and
Shuai Chen}
\authorrunning{Y. Yang, F. Meng et al.}
\institute{School of Information and Communication Engineering \\University of Electronic Science and Technology of China\\ Chengdu, China\\
\email{fmmeng@uestc.edu.cn}\\
}
\maketitle              
\begin{abstract}
Few-shot segmentation segments object regions of new classes with a few of manual annotations. Its key step is to establish the transformation module between support images (annotated images) and query images (unlabeled images), so that the segmentation cues of support images can guide the segmentation of query images. The existing methods form transformation model based on global cues, which however ignores the local cues that are verified in this paper to be very important for the transformation. This paper proposes a new transformation module based on local cues, where the relationship of the local features is used for transformation. 
To enhance the generalization performance of the network, the relationship matrix is calculated in a high-dimensional metric embedding space based on cosine distance. In addition, to handle the challenging mapping problem from the low-level local relationships to high-level semantic cues, we propose to apply generalized inverse matrix of the annotation matrix of support images to transform the relationship matrix linearly, which is non-parametric and class-agnostic. The result by the matrix transformation can be regarded as an attention map with high-level semantic cues, based on which a transformation module can be built simply.
The proposed transformation module is a general module that can be used to replace the transformation module in the existing few-shot segmentation frameworks. We verify the effectiveness of the proposed method on Pascal VOC 2012 dataset. The value of mIoU achieves at 57.0\% in 1-shot and 60.6\% in 5-shot, which outperforms the state-of-the-art method by 1.6\% and 3.5\%, respectively.

\keywords{Few-shot Segmentation \and Transformation Module \and Attention \and Matrix Transformation}
\end{abstract}
\section{Introduction}

Image segmentation is a basic computer vision task\cite{ref_fcn}. In recent years, with the rapid development of deep learning method, several convolution neural network based segmentation methods have improved the performance of image segmentation greatly, such as FCN\cite{ref_fcn} and DeepLab v3\cite{ref_deeplab}. However, these methods rely heavily on a large amount of annotations. In order to overcome this shortcoming, few-shot segmentation task \cite{ref_first_one} is proposed to achieve segmentation of new class with a few of manual annotations, such as one annotation (1-shot segmentation) and five annotations (5-shot segmentation). Few-shot segmentation is a challenging task due to the asymmetry of training data and testing data.

In few-shot segmentation task, the annotated and unlabeled images are called support images and query images respectively\cite{ref_first_one}. The existing models\cite{ref_first_one}\cite{ref_conditional}\cite{ref_sgone}\cite{ref_aaai}\cite{ref_cvpr} usually consist of three terms. 1) The support branch that extracts feature from support images. 2) The query branch that extracts feature from query images. 3) The transformation module that transfers the features between support branch and query branch to facilitate the segmentation of query branch. The essential term is the transformation module and the most challenging obstacle is how to design a transformation module that is class-agnostic, so that the transformation module can be generalized to new classes efficiently. The existing methods\cite{ref_sgone}\cite{ref_cvpr} use the global cues of the support image to model the transformation process, which however ignores the geometry relationships of the local features. This paper demonstrates that the geometry relationships of the local features are very useful to the transformation module.

This paper proposes a new transformation module based on local cues, where the relationship of the local features is used to accomplish the transformation. 
Our idea is to use linear transformation of the relationship matrix in a high-dimensional metric embedding space to accomplish the transformation. 
To this end, we firstly map the local features into an embedding space, where cosine distance is used to obtain the relationship matrix of local features. Then, the relationship matrix is transformed linearly by the generalized inverse matrix of the annotated matrix of support image. After linear transformation, the result is regarded as an attention map containing high-level semantic information, by which we establish a new attention transformation module. We verify the effectiveness of our transformation module on Pascal VOC 2012 dataset\cite{ref_voc}. The value of mIoU achieves at 57.0\% in 1-shot and 60.6\% in 5-shot, which outperform the state-of-the-art method by 1.6\% and 3.5\%, respectively.

\section{Proposed Method}
\subsection{Problem Definition}

Few-shot segmentation is a task that uses a few of annotations to segment unknown images for new classes. Let $S=\{(I^i_s,Y^i_s)\}^k_{i=1}$ be a set of support images and the corresponding manual annotations. Let $Q=\{I_q\}$ be query images set that needs to be segmented. 
The images in $S$ and $Q$ belong to the same new class $l\in\{L_{test}\}$. Let $\{L_{train}\}$ be the training dataset of known classes that already exist, and  $\{L_{train}\}\cap\{L_{test}\}=\emptyset$. The goal of few-shot segmentation is to build a model $f(I_q, S)$ by $\{L_{train}\}$ that outputs binary mask $Y_q$ for query image $I_q$ based on $S$.

\subsection{Overview}
\begin{figure}
\includegraphics[width=\textwidth, height=2.1in]{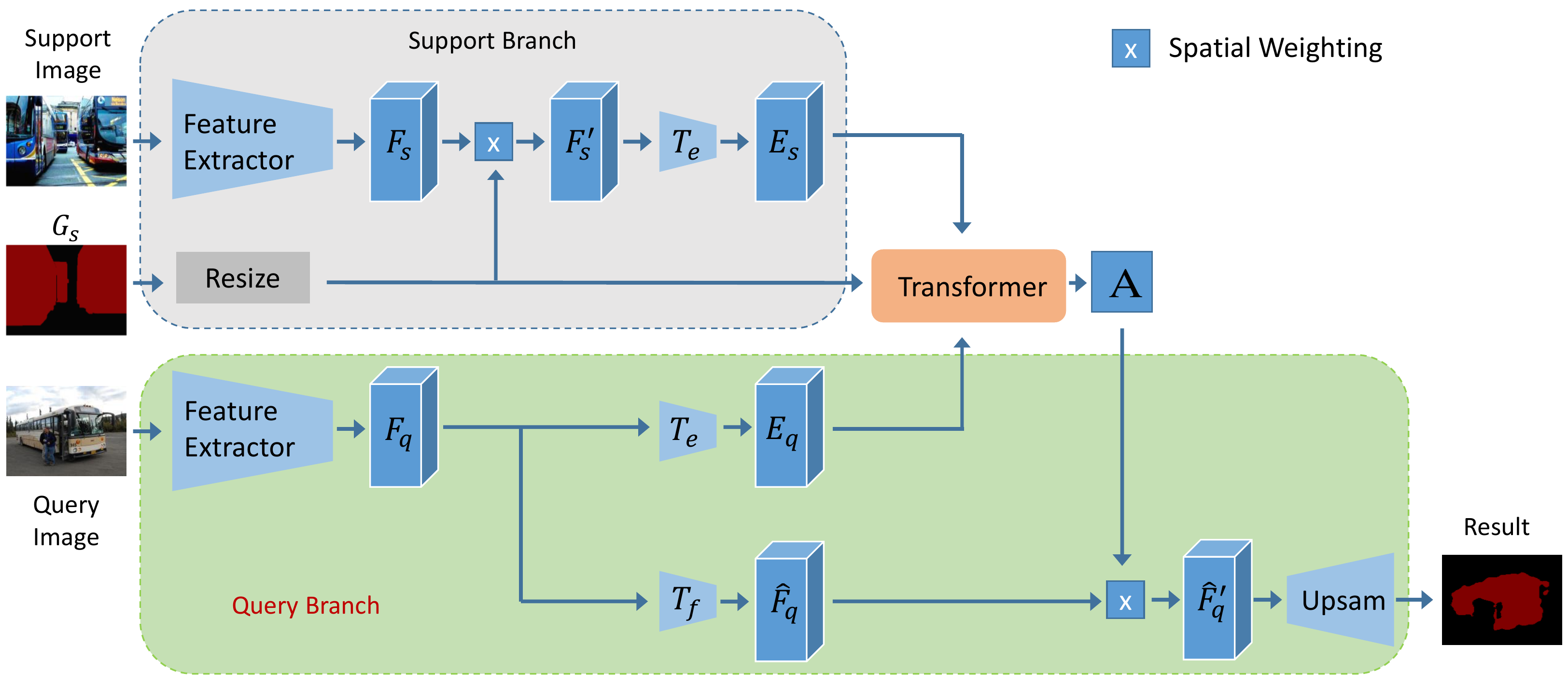}
\caption{The pipeline of the proposed method. The \textit{Feature  Extractor} extracts the features $F_s$ and $F_q$. The feature $F_s$ is weighted spatially by the groundtruth mask $G_s$ to obtain the feature $F_{s}^{'}$. The deep features $F_q$ and $F_{s}^{'}$ are mapped into an embedding space by $T_e$ to get $E_q$ and $E_s$ respectively. Simultaneously, the feature $F_q$ is learned by $T_f$ to get $\hat{F_q}$. Based on $E_s$, $E_q$ and $G_s$, the proposed $Transformer$ outputs attention map $A$, which weights $\hat{F_q}$ spatially. The segmentation result is obtained through $Upsam$ module finally. }\label{fig1}
\end{figure}

Similar to the existing few-shot segmentation network, the proposed framework includes a support branch, a query branch and a transformation module, as shown in Fig. \ref{fig1}. In order to make the network more generalized to unseen classes, our feature extraction backbone adopts relatively shallow layers, such as the first three layers of $Resnet50$\cite{ref_resnet}. In addition, the support branch and the query branch share the feature extraction backbone.

After obtaining deep features $F_s$ and $F_q$ from support image and query image, $F_s$ is weighted spatially by the annotated mask $G_{s}$ to get the features $F^{'}_s$, which guarantees the features $F^{'}_s$ only containing corresponding foreground regions. Such process can be represented as
\begin{equation}
F^{'}_{s}{(i, j)} = F_{s}(i, j) \times G_{s}(i, j) \label{eqution1}
\end{equation}
where $i$, $j$ is spatial location of feature map $F_s$ or annotated mask  $G_s$.

Then, the learned features $F_q$ and $F^{'}_s$ are mapped into a high-dimensional embedding space by further convolution operations $T_e$ to get corresponding embedding features $E_q$ and $E_s$ respectively, so that cosine distance can be used to calculate the relationship between the feature pixels in this space. Simultaneously, the feature $F_q$ is learned by convolution operation $T_f$ to get $\hat{F_q}$. 

Based on the embedding $E_s$, $E_q$ and the groundtruth mask of support images $G_s$, the proposed $Transformer$ applies linear transformation of matrix to obtain the attention map $A$ with high-level semantic cues. The detailed description refers to section \ref{section_tran}.
The attention map $A$ finally filters the deep features $\hat{F_q}$ to $\hat{F_q^{'}}$ by 
\begin{equation}
\hat{F'_q}(i, j) = \hat{F_q}(i, j) \times A(i, j) \label{eqution2}
\end{equation}
where $i$, $j$ is spatial location of feature maps $\hat{F_q}$ or attention map $A$. It is seen that the attention map $A$ indicates a rough object area to be segmented. This coarse object area provides high-level semantic information for the subsequent $Upsam$ sub-network.

\begin{figure}
\includegraphics[width=\textwidth]{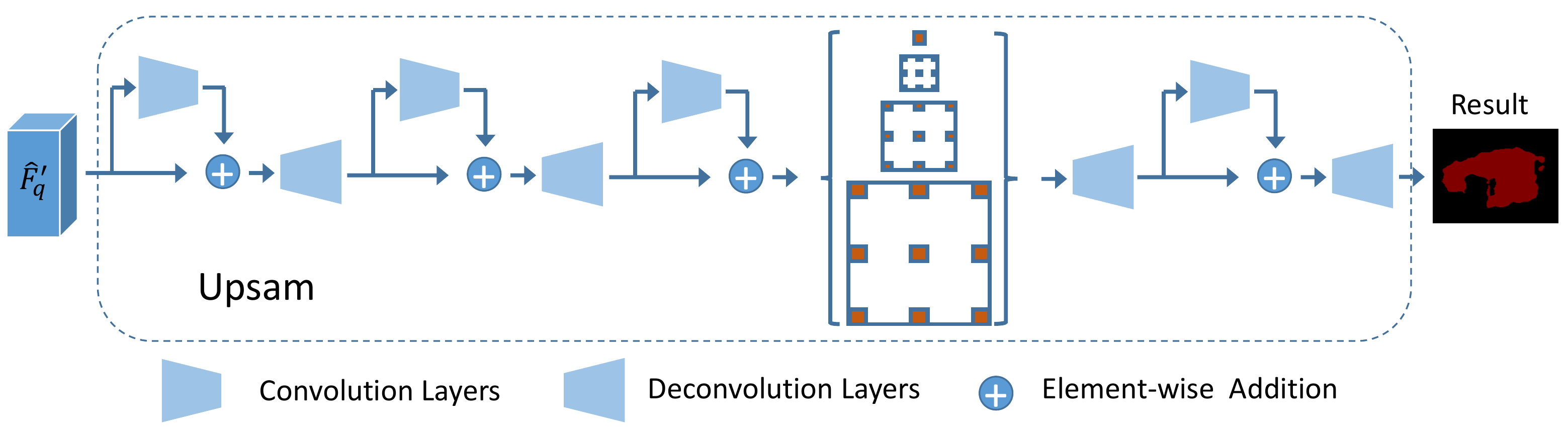}
\caption{The structure of $Upsam$ module. The residual connection\cite{ref_resnet} and multi-scale strategy ASPP\cite{ref_crfv1} are implemented to handle the scale variations of object.} \label{fig3}
\end{figure}

We next use $Upsam$ sub-network to generate segmentation mask from $\hat{F_q}$. The $Upsam$ sub-network is shown in Fig. \ref{fig3}. Specifically, in order to handle the scale changes of the object better, we introduce multi-scale feature fusion module ASPP\cite{ref_crfv1} and residual connection \cite{ref_resnet} in $Upsam$. The output of $Upsam$ is the probability map $M$ with the same size to the query image, and the cross entropy loss in Eq. (\ref{eqution9}) is used to supervise the training of the model, i.e.,

\begin{equation}
L_m = \sum_{i}\sum_{j}-(Y(i, j)log(M(i, j)) + (1-Y(i, j))log(1-M(i, j)))
 \label{eqution9}
\end{equation}
where $Y$ is the groundtruth mask of the query image, $M$ is the predicted probability map of our few-shot segmentation model, and $i$, $j$ is the spatial location of $Y$ or $M$.

\subsection{Transformation module}\label{section_tran}
Existing methods often establish the transformation between support image and query image by the global features of the support image, thus lose local geometric information, which however is also important to the transformation. The proposed transformation module is designed based on the relationship between local pixels (represented by the relationship matrix in NON-Local model\cite{ref_non_local}), and more accurate segmentation can be realized by propagating local relationships. 

\begin{figure}
\includegraphics[width=\textwidth]{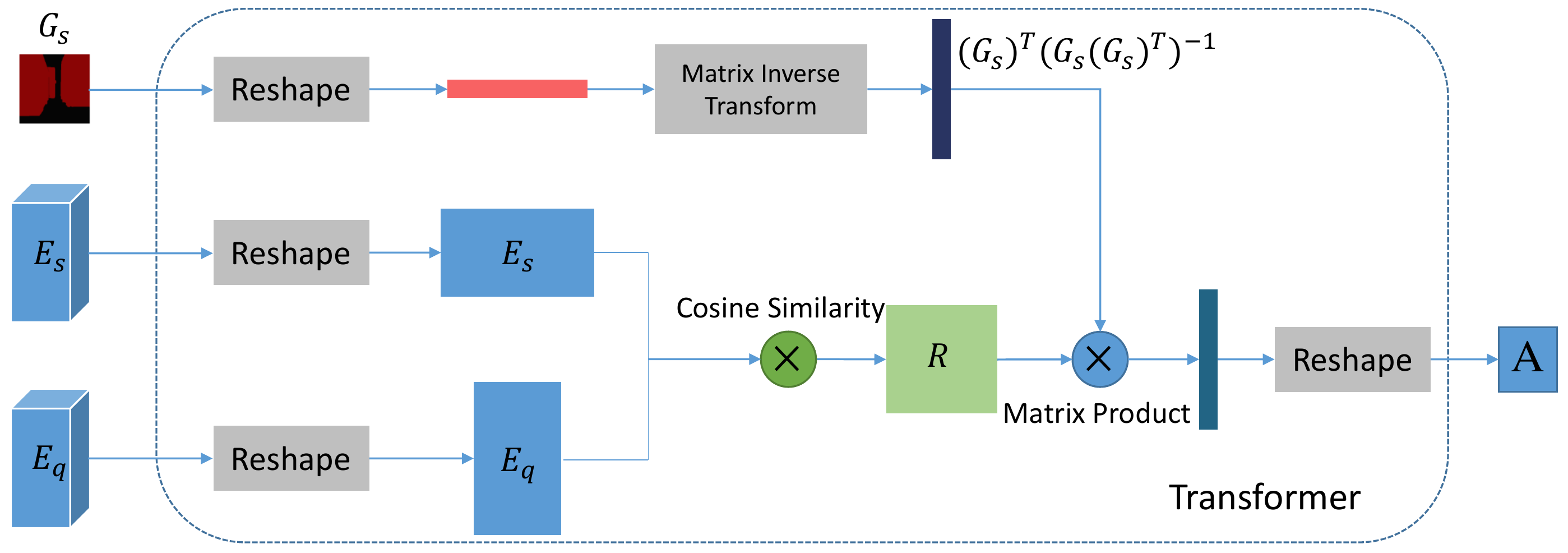}
\caption{The detailed information of the proposed transformation module. The embedding $E_s$ and $E_q$ are reshaped to matrix. Then, the relationship matrix $R$ is obtained based on $E_q$ and $E_s$ by cosine similarity. Finally, the relationship matrix $R$ is transformed linearly by the matrix of generalized inverse matrix of $G_s$. After reshape operator, the attention map $A$ is obtained.} \label{fig2}
\end{figure}

The detailed steps are illustrated in Fig. \ref{fig2}. The transformation is achieved by the linear transformation of the relationship matrix. Specifically, the relationship matrix between $E_s$ (for support image) and $E_q$ (for query image) is linearly transformed by the generalized inverse matrix of $G_s$ (groundtruth mask matrix of the support image), which overcomes the difficulty of transforming local relationship matrix to high-level semantic information. The result of linear transformation is the attention map $A$ of the query image.

\subsubsection{Relationship Matrix}

For few-shot segmentation task, it is very important to model the relationship between each pair of deep local features of support image and query image. Due to the local computational nature of the convolution operation, the relationships between long-distance pixels cannot be established directly. 
Therefore, NON-Local\cite{ref_non_local} structure was proposed to conquer it,
where the feature tensor is reshaped into a matrix, and a relationship matrix is established by matrix product. This relationship matrix contains the relationships between each pair of local deep features.  We imitate the relationship matrix in NON-Local\cite{ref_non_local} to establish the relationship in few-shot segmentation.

In NON-Local\cite{ref_non_local}, it is only desirable to establish long-distance constraints, and the matrix product is just used to describe the relationship between two local features. For few-shot segmentation task, this is a rough description of feature similarity. Therefore, in our proposed transformation module, the feature is firstly mapped into an embedding space, in which the cosine distance can be used to calculate the relationship between local features. Such process is represented as
\begin{equation}
 R_{ij} = \frac{\langle E_{si},E_{qj} \rangle}{{\|E_{si}\|}_2 {\|E_{qj}\|}_2}\label{eqution3}
\end{equation}
where $E_{si}$ represents the $i$th local information in the embedding $E_{s}$, and $E_{qj}$ represents the $j$th local information in the embedding $E_{q}$. So we have established a relationship matrix $R$ between query image and support image.
 
\subsubsection{Linear transformation based on generalized inverse matrix}
With the above relationship matrix $R$, how to convert this relationship matrix $R$ to high-level semantic information of query image becomes another key point. 
Let $G_s$ and $G_q$ be the binary groundtruth mask of support image and query image respectively. Based on Eq. (\ref{eqution3}), the true relationship matrix $R_{truth}$ between query image and support image can be simplified by matrix product between $G_s$ and $G_q$, i.e.,
\begin{equation}
 R_{truth} = G_{q} \cdot G_{s}
\label{eqution4}
\end{equation}
where $\cdot$ demonstrates matrix product. The original size of $G_s$ and $G_q$ are $H \times W$. The reshaped size of $G_s$ and $G_q$ are $1 \times HW$ and $HW \times 1$ respectively. The size of $R_{truth}$ is $HW \times HW$, which contains the relationship information of each pair of local feature pixels of $E_{q}$ and $E_{s}$. Our target is to obtain $G_{q}$ based on Eq. (\ref{eqution4}).

We suppose the matrix $R$ is approximately equal to the true relationship matrix $R_{truth}$ between query image and support image. Furthermore, we relax the binary groundtruth mask $G_q$ to the soft attention map $A$, which provides high-level semantic information. Since $G_{s}$ is known for few-shot segmentation task, the problem is transformed to get the attention map $A$ based on, 
\begin{equation}
 R = A \cdot G_{s}
\label{eqution20}
\end{equation}

Moreover, since the $G_s$ is not square matrix, its inverse matrix does not exist. But it can be regarded as matrix with row full rank. According to the generalized inverse matrix theory \cite{ref_inverse}, the transformation problem can be represented by:
\begin{equation}
A = R \cdot [(G_s)^{T}(G_s(G_s)^{T})^{-1}]
\label{eqution5}
\end{equation}
where $\cdot$ demonstrates matrix product. $(G_s)^{T}(G_s(G_s)^{T})^{-1}$ is the right inverse matrix (one type of generalized inverse matrix) of $G_s$. This is just a process that applies the generalized inverse matrix of $G_s$ to linearly transform the relationship matrix $R$. Finally, the attention map can be obtained by Eq. (\ref{eqution5}) directly.

In order to ensure that the learned relation matrix $R$ is consistent with $R_{truth}$ during training, the mean square error loss is used to supervise it, i.e., 

\begin{equation}
L_{r} = \left \|R-R_{truth} \right \|_{2}^2
\label{eqution6}
\end{equation}

\subsubsection{Attention Map}
By linearly transforming the relationship matrix $R$ by Eq. (\ref{eqution5}), the attention map of the query image $A$ is obtained, with reshaped size to $H \times W$. Moreover, we normalize it to $0 \sim 1$ by
\begin{equation}
\hat{A} = \frac{A-min(A)}{max(A)-min(A)}
\label{eqution7}
\end{equation}
where $\hat{A}$ is normalized counterpart of $A$, for the convenience of expression, we do not distinguish them.

The deep feature of the query image $\hat{F_q}$ is filtered by the normalized attention map $\hat{A}$ to get $\hat{F_q^{'}}$ by Eq. (\ref{eqution2}). Then $\hat{F_q^{'}}$ is proceeded by $Upsam$ (as shown in Fig. \ref{fig3}) to obtain the segmentation result.

In order to ensure the accuracy of the attention map $A$, we regard it as the foreground probability map of the segmentation result, and $1-A$ as the background probability map. The two maps are concatenated and resized to the same size of original query image (by bilinear interpolation). The combined map $M_a$ can be regarded as a segmentation result and supervised by the cross entropy loss.
\begin{equation}
L_a = \sum_{i}\sum_{j}-(Y(i, j)log(M_a(i, j)) + (1-Y(i, j))log(1-M_a(i, j)))
 \label{eqution8}
\end{equation}
where $Y$ is the groundtruth mask of the query image, $M_a$ is the probability map derived from attention map $A$. $i$ and $j$ are the spatial location of $Y$ and $M_a$.

For 5-shot, there are five support images. In order to combine the attention maps provided by the five different images, we simply average the attention maps by
\begin{equation}
A_{5-shot} = \frac{1}{5} \sum_i A_i
 \label{eqution11}
\end{equation}

In the training stage, we combine the three losses in Eq. (\ref{eqution9}), (\ref{eqution8}) and (\ref{eqution6}) to supervise the learning of our model, i.e., 
\begin{equation}
L = \lambda_mL_m + \lambda_aL_a + \lambda_rL_r
 \label{eqution10}
\end{equation}
where $\lambda_m$, $\lambda_a$, $\lambda_r$ are the weights of corresponding loss function.

\section{Experiment}

The project of our method is built based on the Pytorch library, Adam\cite{ref_adam} optimizer is adopted to update the parameters, and all experimental code is executed on a machine equipped with a Titan XP GPU. We set the initial learning rate to 1e-4. The backbone network of our feature extraction is pre-trained on ImageNet \cite{ref_imagenet} dataset, and the parameters of the previous layers of backbone are frozen, and we apply the first three layers of $Resnet50$ \cite{ref_resnet} as our backbone.

\begin{table}[h]
        \centering
        \caption{The detailed setting for splitting the sub-dataset to evaluate the few-shot segmentation. There are 4 sub-datasets, and $PASCAL-5^i$ represents the $i$th subset, where $i=\{0, 1, 2, 3\}$. When the $i$-th sub-dataset is selected for evaluation, the rest three datasets are used for training.}
        \label{table_subdataset}
        \begin{tabular}{|c|c|}
                \hline
                sub-dataset&corresponding classes\\
                \hline
                $PASCAL-5^0$&aeroplane,bicycle,bird,boat,bottle\\
                $PASCAL-5^1$&bus, car, cat, chair, cow\\
                $PASCAL-5^2$&diningtable, dog, horse, motorbike, person\\
                $PASCAL-5^3$&potted plant, sheep, sofa, train, tv/monitor\\
                \hline
        \end{tabular}
\end{table}

\subsection{Detail of Implementation}
We validate the proposed method on the Pascal VOC 2012\cite{ref_voc} dataset and its enhanced dataset SDS\cite{ref_sds}. Similar to the existing methods \cite{ref_first_one,ref_conditional,ref_sgone,ref_aaai,ref_cvpr}, we split images of 20 classes into four subsets, each of which contains images of five classes, the detailed description can be found in Table \ref{table_subdataset}. For these four subsets, three of them are selected as the training set, and the rest one is used as the test set to validate the effectiveness of the proposed method. In the training stage, we randomly select two images for each class, one as a support image and another as a query image until all images of training classes were selected. In the testing stage, in order to make a fair comparison with the existing methods, we use the same random seed in the existing method to sample the same 1000 pairs of images as the test data for each evaluation sub-dataset.

\begin{table}[htp]
        \centering
        \caption{The comparison results (mIoU value) on four evaluation sub-datasets in 1-shot. The best results are in bold.}
        \label{table_1shot_miou}
        \begin{tabular}{|c|cccc|c|}
                \hline
               Methods&$PASCAL-5^0$&$PASCAL-5^1$&$PASCAL-5^2$&$PASCAL-5^3$&Mean\\
               \hline
                1-NN & 25.3& 44.9& 41.7& 18.4& 32.6 \\
                LogReg& 26.9& 42.9& 37.1& 18.4& 31.4\\
                Siamese& 28.1& 39.9& 31.8& 25.8& 31.4\\
                OSVOS\cite{ref_vdieo}& 24.9& 38.8& 36.5& 30.1& 32.6\\
                OSLSM\cite{ref_first_one}& 33.6& 55.3& 40.9& 33.5& 40.8\\
                co-FCN\cite{ref_conditional}& 36.7& 50.6& 44.9& 32.4& 41.1\\
                SG-One\cite{ref_sgone}&40.2 &58.4 &48.4 &38.4 &46.3\\
                CA-Net\cite{ref_cvpr}&52.5 &65.9 &51.3 &51.9 &55.4\\
                Ours& \bfseries{52.8}& \bfseries{69.6}& \bfseries{53.2}& \bfseries{52.3}& \bfseries{57.0}\\
                \hline
        \end{tabular}
\end{table}

\begin{table}[htp]
        \centering
        \caption{The comparison results (mIoU value) on four evaluation sub-datasets in 5-shot. The best results are in bold.}
        \label{table_5shot_miou}
        \begin{tabular}{|c|cccc|c|}
                \hline
               
               Methods&$PASCAL-5^0$&$PASCAL-5^1$&$PASCAL-5^2$&$PASCAL-5^3$&Mean\\
               \hline
                1-NN & 34.5& 53.0& 46.9& 25.6& 40.0 \\
                LogReg& 25.9& 51.6& 44.5& 25.6& 39.3\\
                Co-segmentation\cite{ref_coseg}& 25.1& 28.9& 27.7& 26.3& 27.1\\
               
                OSLSM\cite{ref_first_one}& 35.9& 58.1& 42.7& 39.1& 43.9\\
                co-FCN\cite{ref_conditional}& 37.5& 50.0& 44.1& 33.9& 41.4\\
                SG-One\cite{ref_sgone}&41.9 &58.6 &48.6 &39.4 &47.1\\
                CA-Net\cite{ref_cvpr}&55.5 &67.8 &51.9 &53.2 &57.1\\
                Ours& \bfseries{57.9}& \bfseries{69.9}& \bfseries{56.9}& \bfseries{57.5}& \bfseries{60.6}\\
                \hline
        \end{tabular}
\end{table}

 \begin{figure}[htp]
\includegraphics[width=\textwidth, height=10.5cm]{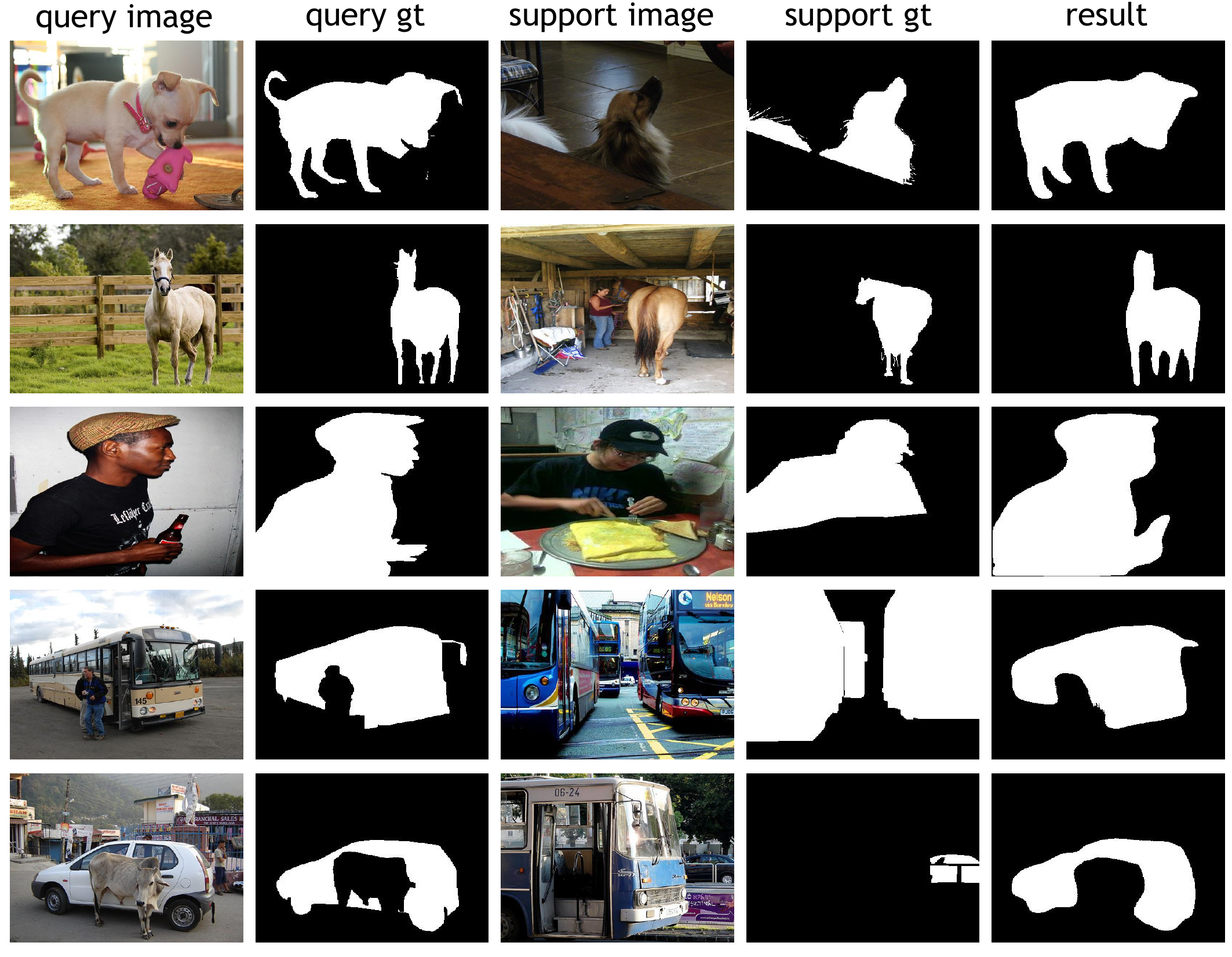}
\caption{The subjective results of the proposed method. From left to right: query image, ground-truth mask of the query image, the support image, ground-truth mask of the support image and the segmentation result, respectively.} \label{fig4}
\end{figure}
\begin{table}[htp]
        \centering
        \caption{The comparison results (FB-IoU value) on four evaluation sub-datasets in 1-shot and 5-shot. The best results are in bold.}
        \label{table_1shot_FB_IoU}
        \begin{tabular}{|c|ccccccc|}
                \hline
                Methods  & OSLSM\cite{ref_first_one}&  co-FCN\cite{ref_conditional}&  PL\cite{ref_pl}& 
                SG-One\cite{ref_sgone}&
                A-MCG\cite{ref_aaai}  & CA-Net\cite{ref_cvpr}& Ours\\
                \hline
                1-shot& 61.3& 60.1& 61.2& 63.1& 61.2 &66.2& \bfseries{71.8}\\
                \hline
                5-shot& 61.5& 60.2& 62.3& 65.9& 62.2& 69.6& \bfseries{74.6}\\
                \hline
        \end{tabular}
\end{table}

\begin{table}[htp]
        \centering
        \caption{The ablation results of three loss functions. The ticking indicates that the loss function is used.}
        \label{table_ablation_loss}
        \begin{tabular}{|c|p{10mm}p{10mm}p{10mm}|c|c|}
                \hline
                k-shot&$L_m$&$L_r$&$L_a$&mIoU&FB-IoU\\
                \hline
                1-shot& \checkmark & & &55.3&70.2\\
                \hline
                1-shot& \checkmark &\checkmark & &55.9&70.9\\
                \hline
                 1-shot& \checkmark &&\checkmark &56.0&71.1\\
                \hline
                 1-shot& \checkmark & \checkmark&\checkmark &57.0&71.8\\
                \hline
        \end{tabular}
\end{table}
 
 We use the mean intersection over union of foreground (mIoU) to measure the performance of our proposed method, which is widely used in few-shot segmentation. In addition, the FB-IoU proposed in co-FCN\cite{ref_conditional} is also considered, which includes mean intersection over union of foreground and background.
 
\subsection{Comparison with Benchmarks}
 In order to verify the effectiveness of our method, we compare with existing method in 1-shot and 5-shot. We follow CA-Net\cite{ref_cvpr} to adopt DenseCRF\cite{ref_crf} and multi-scale evaluation strategy to improve the performance, which are always employed in existing\cite{ref_crfv1} semantic segmentation method. The detailed results can be found in Table \ref{table_1shot_miou}, Table \ref{table_5shot_miou} and Table \ref{table_1shot_FB_IoU}. We can see the values of mIoU by our method achieve at 57.0\% in 1-shot and 60.6\% in 5-shot, which outperform the state-of-the-art few-shot segmentation method CA-Net\cite{ref_cvpr} by 1.6\% and 3.5\% respectively. The improvement in 5-shot indicates the superiority of our method when it comes to more annotations. In addition, the values of FB-IoU in 1-shot and 5-shot achieve at 71.8\%, 74.6\% respectively, which also outperforms the comparison methods obviously.
 
\begin{table}[htp]
        \centering
        \caption{The effectiveness of Dense-CRF post-processing and Multi-scale strategy are demonstrated. The ticking indicates that the strategy is used.}
        \label{table_ablation_crf}
        \begin{tabular}{|c|cc|c|c|}
                \hline
                k-shot &Dense-CRF & Multi-scale& mIoU &FB-IoU\\
                \hline
                1-shot&  &&56.1&71.0\\
                \hline
                1-shot& \checkmark &&56.7&71.5\\
                \hline
                1-shot& &\checkmark &56.6&71.3\\
                \hline
                1-shot& \checkmark &\checkmark&57.0&71.8\\
                \hline
        \end{tabular}
\end{table}
\subsection{Ablation}
 In the training stage, the weight $\lambda_m$, $\lambda_a$ and $\lambda_r$ are set to 1. In order to validate the effectiveness of our three loss functions, ablation experiment is implemented. The detailed results can be found in Table \ref{table_ablation_loss}. We can see $L_a$ and $L_r$ improve the performance by 1.1\% and 1.0\% respectively.
 In addition, we follow CA-Net\cite{ref_cvpr} to employ DenseCRF\cite{ref_crf} and multi-scale evaluation in our test stage. The ablation of these two strategies is also conducted, and the detailed results can be found in Table \ref{table_ablation_crf}. We can see that DenseCRF\cite{ref_crf} and the multi-scale evaluation strategy can improve the mIoU value by 0.4\% and 0.3\% respectively.

\subsection{Subjective Result}
The subjective results of the proposed method are shown in Fig. \ref{fig4}. The support image, the ground-truth mask of the support image, the query image, the ground-truth mask of the query image and the segmentation result are displayed from left column to right column, respectively.  It is seen that the proposed method segments objects from these images successfully.

\section{Conclusion}
This paper proposes a new transformation module for few-shot segmentation. Rather than focusing on global cues, the relationships of local features are used to form the transformation. Local feature relationship matrix calculated by the cosine similarity is used to represent the relationships of local features. Linear transformation of relationship matrix based on generalized inverse of the groundtruth matrix is implemented to transform the relationship matrix. We also map the features into a high-dimensional metric embedding space to enhance the generalization of the proposed module. We propose a new few-shot segmentation network based on transformation module, and better results are obtained in terms of both mIoU value and FB-IoU value.

\section*{Acknowledgment}
This work was supported in part by the National Natural Science Foundation of China under Grant 61871087, Grant 61502084, Grant 61831005, and Grant 61601102, and supported in part by Sichuan Science and Technology Program under Grant 2018JY0141.

\end{document}